\documentclass[11pt,a4paper]{article}
\usepackage{times,latexsym}
\usepackage{url}
\usepackage[T1]{fontenc}

\usepackage[acceptedWithA]{tacl2018v2}

\usepackage{xspace,mfirstuc,tabulary}

\newif\iftaclinstructions
\taclinstructionsfalse 
\iftaclinstructions
\renewcommand{\confidential}{}
\renewcommand{\anonsubtext}{(No author info supplied here, for consistency with
TACL-submission anonymization requirements)}
\newcommand{\instr}
\fi

\usepackage{microtype}
\usepackage{graphicx}
\usepackage{multirow}
\usepackage{multicol}
\usepackage{makecell}
\usepackage{pifont}
\usepackage{array}
\newcolumntype{L}[1]{>{\raggedright\let\newline\\\arraybackslash\hspace{0pt}}m{#1}}
\newcolumntype{C}[1]{>{\centering\let\newline\\\arraybackslash\hspace{0pt}}m{#1}}
\newcolumntype{R}[1]{>{\raggedleft\let\newline\\\arraybackslash\hspace{0pt}}m{#1}}
\newcolumntype{B}[1]{>{\raggedright\let\newline\\\arraybackslash\hspace{0pt}}p{#1}}
\newcolumntype{N}[1]{>{\centering\let\newline\\\arraybackslash\hspace{0pt}}p{#1}}
\newcolumntype{M}[1]{>{\raggedleft\let\newline\\\arraybackslash\hspace{0pt}}p{#1}}

\usepackage{soul}

\newcommand{\ebhd}{explanation-based human debugging\xspace} 
 
\newcommand{\EBHD}{EBHD\xspace} 
\newcommand{\numpapers}{15\xspace} 

\title{Explanation-Based Human Debugging of NLP Models: A Survey}

\author{Piyawat Lertvittayakumjorn {\normalfont and} Francesca Toni\\
  Department of Computing \\
  Imperial College London, UK\\
  \texttt{\{pl1515, ft\}@imperial.ac.uk}}

\date{}

\begin{document}
\maketitle
\begin{abstract}
Debugging a machine learning model is hard since the bug usually involves the training data and the learning process. This becomes even harder for an opaque deep learning model if we  have no clue about how the model actually works.
In this survey, we review papers that exploit explanations to enable humans to give feedback and debug NLP models. We call this problem \emph{\ebhd (\EBHD)}.
In particular, we categorize and discuss existing work along three dimensions of \EBHD (the bug context, the workflow, and the experimental setting), compile findings on how \EBHD components affect the feedback providers, and highlight open problems that could be future research directions.
\end{abstract}

\section{Introduction}
Explainable AI focuses on generating explanations for AI models as well as for their predictions. 
It is gaining more and more attention these days since explanations are necessary in many 
applications, especially in high-stake domains such as healthcare, law, transportation, and finance \cite{adadi2018peeking}. 
Some researchers have explored various merits of explanations to humans, such as 
supporting human decision making \cite{lai2019human,lertvittayakumjorn-etal-2021-supporting},
increasing human trust in AI \cite{jacovi2020formalizing}
and even teaching humans to perform challenging tasks \cite{lai2020chicago}.
On the other hand, explanations can benefit the AI systems as well, e.g.,
when explanations are used
to promote system acceptance \cite{cramer2008effects},
to verify the model reasoning \cite{caruana2015intelligible},
and to find potential causes of errors \cite{han-etal-2020-explaining}. 

In this paper, we review progress to date  
specifically on how explanations have been used in the literature to enable humans to fix bugs in NLP models. We refer to this research area as \emph{\ebhd (\EBHD)}, as a general umbrella term encompassing  \emph{explanatory debugging} \cite{kulesza2010explanatory} and \emph{human-in-the-loop debugging} \cite{lertvittayakumjorn-etal-2020-find}.
We define \EBHD as the process of fixing or mitigating bugs in a trained model using human feedback given in response to explanations for the model.
\EBHD is helpful when the training data at hand leads to suboptimal models (due, for instance, to biases 
or artifacts in the data), and hence human knowledge is needed to verify and improve the trained models.  
In fact, \EBHD is related to three challenging and intertwined issues in NLP: explainability \cite{danilevsky-etal-2020-survey}, interactive and human-in-the-loop learning \cite{amershi2014power,wang2021putting}, and knowledge integration \cite{von2021informed,kim2021knowledge}. 
Although there are overviews for each of these topics
(as cited above), our paper is the first to draw connections among the three towards the final application of model debugging in NLP.

\begin{figure*}[t] 
    \centering
    \includegraphics[width=0.90\linewidth]{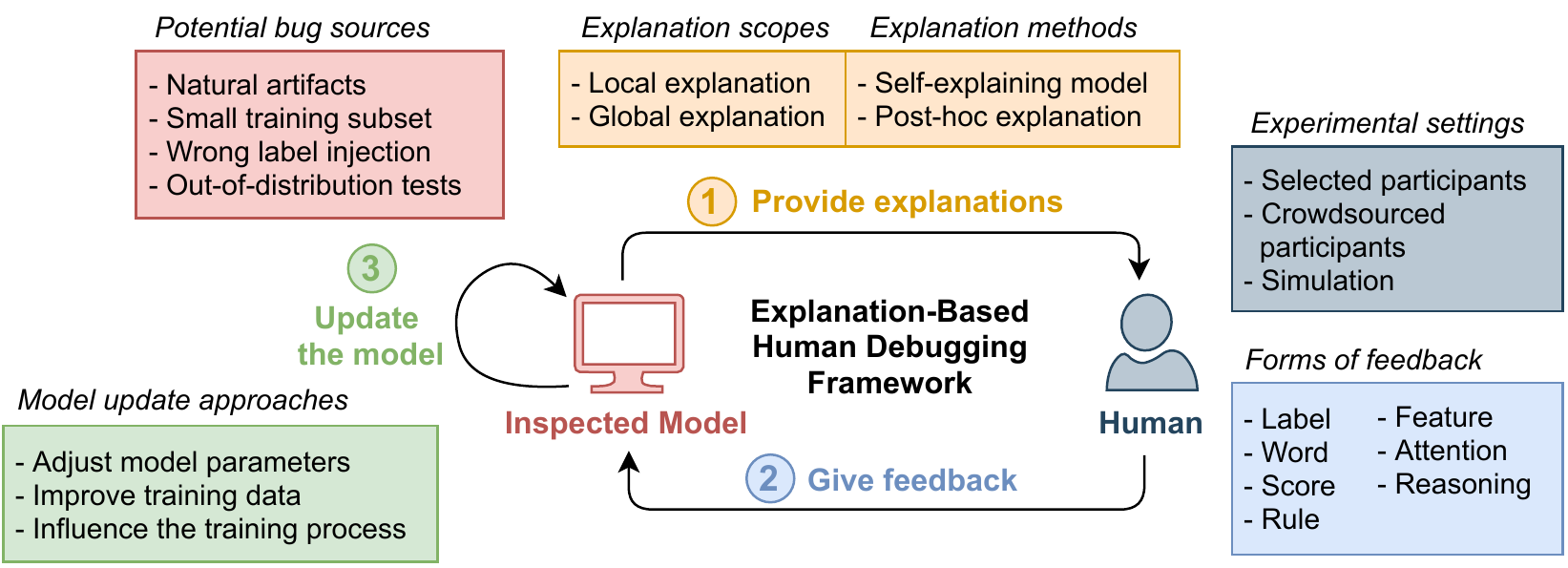}
    \caption{A general framework for \ebhd (\EBHD) of NLP models, consisting of the inspected (potentially buggy) model, the humans providing feedback, and a three-step workflow. Boxes list examples of the options (considered in the selected studies) for the components or steps in the general framework.} \label{fig:overview}
\end{figure*}

\begin{figure*}[t] 
    \centering
    \includegraphics[width=0.90\linewidth]{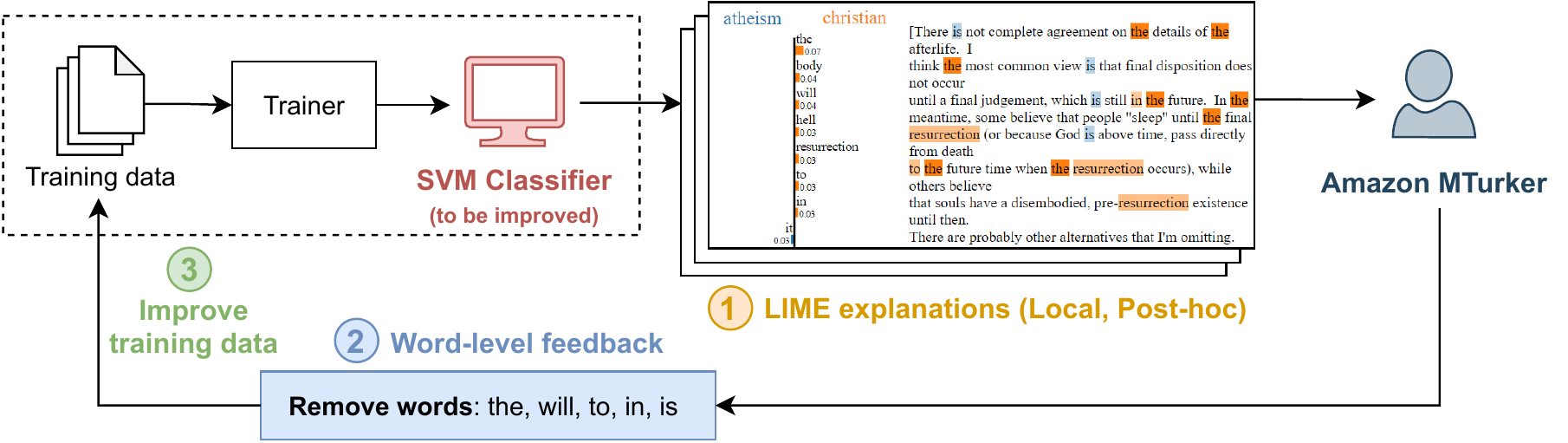}
    \caption{The proposal by \citet{ribeiro2016should} as an instance of the general \EBHD framework.} \label{fig:lime}
\end{figure*}

Whereas most people agree on the meaning of the term \emph{bug} in software engineering, various meanings have been ascribed to this term in machine learning (ML) research.
For example, \citet{selsam2017developing} considered bugs as implementation errors, similar to software bugs, while
\citet{cadamuro2016debugging} defined a bug as a particularly damaging or inexplicable test error.
In this paper, we follow the definition of (model) bugs from \citet{adebayo2020debugging} as contamination in the learning and/or prediction pipeline that makes the model produce incorrect predictions or learn error-causing associations. Examples of bugs include spurious correlation, labelling errors, and undesirable behavior in out-of-distribution (OOD) testing.

The term \emph{debugging} is also interpreted differently by different researchers. 
Some consider debugging as a process of identifying or uncovering causes of model errors \cite{parikh2011human,gralinski-etal-2019-geval}, while others stress that debugging must not only reveal the causes of problems but also fix or mitigate them \cite{kulesza2015principles,yousefzadeh2019debugging}. In this paper, we adopt the latter interpretation.

\paragraph{Scope of the survey.} We focus on work 
using
explanations of NLP models to expose whether there are bugs
and exploit human feedback to fix the bugs (if any). 
To collect relevant papers, we started from some pivotal \EBHD work, e.g., \cite{kulesza2015principles,ribeiro2016should,teso2019explanatory}, and added \EBHD papers citing or being cited by the pivotal work, e.g., \cite{stumpf2009interacting,kulesza2010explanatory,lertvittayakumjorn-etal-2020-find,yao2021refining}.
Next, to ensure that we did not miss any important work, we searched for papers on Semantic Scholar\footnote{\url{https://www.semanticscholar.org/}} using the Cartesian product of five keyword sets: \{debugging\}, \{text, NLP\}, \{human, user, interactive, feedback\}, \{explanation, explanatory\}, and \{learning\}.
With 16 queries in total, we collected the top 100 papers (ranked by relevancy) for each query and kept only the ones appearing in at least 2 out of the 16 query results.
This resulted in 234 papers which we then manually checked, leading to selecting a few additional papers, including \cite{han2020model,zylberajch-etal-2021-hildif}. 
The overall process resulted in \numpapers papers in Table~\ref{tab:papers} as the \textit{selected studies} primarily discussed in this survey. 
In contrast, 
some papers from the following categories appeared in the search results, but were not selected
since, strictly speaking, they are not in the main scope of this survey:
debugging without explanations \cite{kang2018model},
debugging outside the NLP domain \cite{ghai2021explainable,popordanoska2020machine,bekkemoen2021correcting}, 
refining the ML pipeline instead of the model \cite{lourencco2020bugdoc,schoop2020scram}, 
improving the explanations instead of the model \cite{ming2019interpretable}, 
and
work centered on revealing but not fixing bugs \cite{ribeiro-etal-2020-beyond,krause2016interacting,krishnan2017palm}.

\paragraph{General Framework.} 

\EBHD 
consists of
three main 
steps as shown in Figure~\ref{fig:overview}.
First, the explanations, which provide interpretable insights into the inspected model and possibly reveal bugs, are 
given to humans.
Then, the humans inspect the explanations and give feedback in response.
Finally, the feedback is used to update and improve the model.
These steps can be carried out once, as a one-off improvement, or iteratively, depending on how the debugging framework is designed. 

As a concrete example, Figure~\ref{fig:lime} illustrates how \citet{ribeiro2016should} improved an SVM text classifier trained on the 20Newsgroups dataset \cite{Lang95}.
This dataset has many
artifacts which could make the model rely on wrong words or tokens when making predictions, reducing its generalizability\footnote{For more details, please see section~\ref{subsec:bugsources}.}. 
To perform \EBHD, \citet{ribeiro2016should} recruited humans 
from a crowdsourcing platform (i.e., Amazon Mechanical Turk) and asked them to inspect LIME explanations\footnote{LIME stands for Local Interpretable Model agnostic Explanations \cite{ribeiro2016should}. For each model prediction, it returns relevance scores for words in the input text to show how important each word is for the prediction.} (i.e., word relevance scores) for model predictions of ten examples.
Then, the humans gave feedback by identifying words in the explanations that should not have got high relevance scores (i.e., supposed to be the artifacts). These words were then removed from the training data, and the model was retrained. The process was repeated for three rounds, and the results show that the model generalized better after every round. 
Using the general framework in Figure~\ref{fig:overview}, we can break the framework of \citet{ribeiro2016should} into components 
as depicted in Figure~\ref{fig:lime}.  
Throughout the paper, when reviewing 
the selected studies,
we will use the general framework in Figure~\ref{fig:overview} for analysis, comparison, and discussion.

\paragraph{Human Roles.}
To avoid confusion, it is worth noting that there are actually two human roles in the \EBHD process. One, of course, is that of 
\textit{feedback provider(s)}, looking at the explanations and providing feedback (noted as `Human' in Figure~\ref{fig:overview}). 
The other is
that of \textit{model developer(s)}, training the model and organizing the \EBHD process (not shown in Figure~\ref{fig:overview}). 
In practice, a person could be both 
model developer and 
feedback provider. This usually happens during the model validation and improvement phase where the developers try to fix the bugs themselves.
Sometimes, however, other stakeholders could also take the feedback provider role.
For instance, if the model is trained to classify electronic medical records, the developers (who are mostly ML experts) hardly have the medical knowledge to provide feedback. So, they may ask doctors acting as consultants to the development team to be the feedback providers during the model improvement phase.
Further, \EBHD can be carried out after deployment, with end users as the feedback providers.
For example, a model auto-suggesting the categories of new emails to end users can provide explanations supporting the suggestions as part of its normal operation. Also, it can allow the users to provide feedback to both the suggestions and the explanations. 
Then, a routine written by the developers will be triggered to process the feedback and update the model automatically to complete the \EBHD workflow.
In this case, we need to care about the trust, frustration, and expectation of the end users while and after they give feedback.
In conclusion, \EBHD can take place practically both before and after the model is deployed, and many stakeholders can act as the feedback providers, including, but not limited to, the model developers, the domain experts, and the end users.

\paragraph{Paper Organization.}
Section~\ref{sec:categorization} explains the choices made by existing work to achieve \EBHD of NLP models. This illustrates the current state of the field with the strengths and limitations of existing work.   
Naturally, though, a successful \EBHD framework cannot neglect the ``imperfect'' nature of feedback providers, who may not be an ideal oracle.
Hence, Section~\ref{sec:human_factor} compiles relevant human factors which could affect the effectiveness of the debugging process as well as the satisfaction of the feedback providers.
After that, we identify open challenges of \EBHD for NLP in Section~\ref{sec:open_problems} before concluding the paper in Section~\ref{sec:conclusion}.

\section{Categorization of Existing Work} \label{sec:categorization}

\begin{table*}[t]
\setlength{\tabcolsep}{4.5pt}
\centering
\small
\begin{tabular}{|L{0.28\textwidth}| C{0.05\textwidth} C{0.07\textwidth} C{0.10\textwidth}| C{0.05\textwidth} C{0.06\textwidth} C{0.08\textwidth} C{0.05\textwidth} | C{0.07\textwidth}|} 
 \hline
 \multirow{2}{*}{Paper} & \multicolumn{3}{c|}{Context} & \multicolumn{4}{c|}{Workflow} & \multirow{2}{*}{Setting}\\ \cline{2-8}
  & Task & Model & Bug sources & Exp. scope & Exp. method & Feedback & Update & \\
 \hline
 \citet{kulesza2009fixing}&TC&NB&AR&G,L&SE&LB,WS&M,D&SP\\
 \citet{stumpf2009interacting}&TC&NB&SS&L&SE&WO&T&SP\\
 \citet{kulesza2010explanatory}&TC&NB&SS&G,L&SE&WO,LB&M,D&SP\\
 \citet{kulesza2015principles}&TC&NB&AR&G,L&SE&WO,WS&M&SP\\
 \citet{ribeiro2016should}&TC&SVM&AR&L&PH&WO&D&CS\\
 \citet{koh2017understanding}&TC&LR&WL&L&PH&LB&D&SM\\
 \multirow{2}{*}{\citet{ribeiro-etal-2018-semantically}}&VQA&TellQA&AR&\multirow{2}{*}{G}&\multirow{2}{*}{PH}&\multirow{2}{*}{RU}&\multirow{2}{*}{D}&\multirow{2}{*}{SP}\\
 &TC&fastText&AR,OD&&&&&\\
 \citet{teso2019explanatory}&TC&LR&AR&L&PH&WO&D&SM\\
 \citet{cho2019explanatory}&TQA&NeOp&AR&L&SE&AT&T&NR\\
 \citet{khanna2019interpreting}&TC&LR&WL&L&PH&LB&D&SM\\
 \citet{lertvittayakumjorn-etal-2020-find}&TC&CNN&AR,SS,OD&G&PH&FE&T&CS\\
 \citet{smith2020no}&TC&NB&AR,SS&L&SE&LB,WO&M,D&CS\\
 \citet{han2020model}&TC&LR&WL&L&PH&LB&D&SM\\
 \citet{yao2021refining}&TC&BERT*&AR,OD&L&PH&RE&D,T&SP\\
 \citet{zylberajch-etal-2021-hildif}&NLI&BERT&AR&L&PH&ES&D&SP\\
 \hline
 \end{tabular}
\caption{Overview of existing work on \EBHD of NLP models. We use abbreviations as follows: \textbf{Task}: TC = Text Classification (single input), VQA = Visual Question Answering, TQA = Table Question Answering, NLI = Natural Language Inference / 
\textbf{Model}: NB = Naive Bayes, SVM = Support Vector Machines, LR = Logistic Regression, TellQA = Telling QA, NeOp = Neural Operator, CNN = Convolutional Neural Networks, BERT* = BERT and RoBERTa /
\textbf{Bug sources}: AR = Natural artifacts, SS = Small training subset, WL = Wrong label injection, OD = Out-of-distribution tests /
\textbf{Exp. scope}: G = Global explanations, L = Local explanations /
\textbf{Exp. method}: SE = Self-explaining, PH = Post-hoc method /
\textbf{Feedback} (form): LB = Label, WO = Word(s), WS = Word(s) Score, ES = Example Score, FE = Feature, RU = Rule, AT = Attention, RE = Reasoning /
\textbf{Update}: M = Adjust the model parameters, D = Improve the training data, T = Influence the training process /
\textbf{Setting}: SP = Selected participants, CS = Crowdsourced participants, SM = Simulation, NR = Not reported.
} \label{tab:papers}
\end{table*} 

Table~\ref{tab:papers} summarizes the selected studies
along three dimensions, amounting to 
the debugging context (i.e., tasks, models, and bug sources), the workflow (i.e., the three steps in our general framework), and the experimental setting (i.e., the mode of human engagement).
We will discuss these dimensions with respect to the broader knowledge of explainable NLP and human-in-the-loop learning, to shed light on the current state of \EBHD of NLP models.

\subsection{Context} \label{subsec:context}
To demonstrate the debugging process, existing works need to set up the bug situation they aim to fix, including the target NLP task, 
the inspected ML model,
and the source of the bug to be addressed. 

\subsubsection{Tasks} 
Most papers in Table~\ref{tab:papers} focus on text classification with single input (TC) for 
a variety of specific problems such as e-mail categorization \cite{stumpf2009interacting}, topic classification \cite{kulesza2015principles,teso2019explanatory}, spam classification \cite{koh2017understanding}, sentiment analysis \cite{ribeiro-etal-2018-semantically}
and auto-coding of transcripts \cite{kulesza2010explanatory}.
By contrast, \citet{zylberajch-etal-2021-hildif} targeted natural language inference (NLI) which is a type of text-pair classification, predicting whether a given premise entails a given hypothesis.
Finally, two papers involve question answering (QA), i.e., \citet{ribeiro-etal-2018-semantically} (focusing on visual question answering (VQA)) and \citet{cho2019explanatory}
(focusing on table question answering (TQA)).

\citet{ghai2021explainable} suggested that most researchers work on TC because, for this task, it is much easier for lay 
participants to understand explanations and give feedback (e.g., which keywords should be added or removed from the list of top features)\footnote{Nevertheless, some specific TC tasks, such as authorship attribution \cite{juola2007future} and deceptive review detection \cite{lai2020chicago}, are exceptions because lay people are generally not good at these tasks. Thus, they are not suitable for \EBHD.}.
Meanwhile, some other NLP tasks require the feedback providers
to have linguistic knowledge such as part-of-speech tagging, parsing, and machine translation. 
The need for linguists or experts renders experiments for these tasks more difficult and costly.
However, we suggest that there are several tasks 
where the trained models are prone to be buggy but the tasks are underexplored in the \EBHD setting, though they are not too difficult to experiment on with lay people.
\textit{NLI}, the focus of \cite{zylberajch-etal-2021-hildif}, is one of them. Indeed,
\citet{mccoy-etal-2019-right} and \citet{gururangan-etal-2018-annotation} showed that NLI models can exploit annotation artifacts and fallible syntactic heuristics to make predictions rather than learning the logic of the actual task.
Other tasks and their bugs 
include:
\textit{QA}, where \citet{ribeiro-etal-2019-red} found that the answers from models 
are sometimes inconsistent (i.e., contradicting previous answers); and \textit{reading comprehension}, where 
\citet{jia-liang-2017-adversarial} showed that 
models, which answer a question by reading a given paragraph, can be fooled by an irrelevant sentence 
being appended to the paragraph. 
These non-TC NLP tasks would be worth exploring further in the \EBHD setting.

\subsubsection{Models} 
Early work used Naive Bayes models with bag-of-words (NB) as text classifiers \cite{kulesza2009fixing,kulesza2010explanatory,stumpf2009interacting}, which are relatively easy to generate explanations for and to incorporate human feedback into (discussed in section~\ref{subsec:workflow}).
Other traditional models used include logistic regression (LR) \cite{teso2019explanatory,han2020model} 
and support vector machines (SVM) \cite{ribeiro2016should}, both with bag-of-words features.
The next generation of tested models involves word embeddings.
For text classification, \citet{lertvittayakumjorn-etal-2020-find} 
focused on convolutional neural networks (CNN) \cite{kim-2014-convolutional} and touched upon bidirectional LSTM networks \cite{hochreiter1997long}, while \citet{ribeiro-etal-2018-semantically} used fastText, relying also on n-gram features \cite{joulin-etal-2017-bag}.
For VQA and TQA, the inspected models used attention mechanisms for attending to relevant parts of the input image or table. These models are Telling QA \cite{zhu2016visual7w} and Neural Operator (NeOp) \cite{pmlr-v95-cho18a}, used by \citet{ribeiro-etal-2018-semantically} and \citet{cho2019explanatory}, respectively.
While the NLP community nowadays is mainly driven by pre-trained language models \cite{qiu2020pre}
with many papers studying their behaviors \cite{rogers2021primer,hoover-etal-2020-exbert}, 
only \citet{zylberajch-etal-2021-hildif} and \citet{yao2021refining} have used pre-trained language models, including BERT \cite{devlin-etal-2019-bert} and RoBERTa \cite{liu2019roberta}, as test beds for \EBHD.

\subsubsection{Bug Sources} \label{subsec:bugsources}
Most of the papers in Table~\ref{tab:papers} experimented on training datasets with natural artifacts (AR), which cause spurious correlation bugs (i.e., the input texts having signals which are correlated to but not the reasons for specific outputs) and undermine models' generalizability.
Out of the \numpapers papers we surveyed, 5 used the 20Newsgroups dataset \cite{Lang95} as a case study, since it has lots of natural artifacts. 
For example, some punctuation marks appear more often in one class due to the writing styles of the authors contributing to the class, so the model uses these punctuation marks as clues to make predictions. However, because 20Newsgroups is a topic classification dataset, a better model should focus more on the topic of the content since the punctuation marks can also appear in other classes, especially when we apply the model to texts in the wild.
Apart from classification performance drops, natural artifacts can also cause model biases, as shown in \cite{dearteaga2019biosbias,park-etal-2018-reducing} and debugged in \cite{lertvittayakumjorn-etal-2020-find,yao2021refining}.

In the absence of strong natural artifacts, bugs can still be simulated using several techniques. 
First, using only a small subset of labeled data (SS) for training could cause the model to exploit spurious correlation leading to poor performance \cite{kulesza2010explanatory}. 
Second, injecting wrong labels (WL) into the training data can obviously blunt the model quality \cite{koh2017understanding}. 
Third, using out-of-distribution tests (OD) can reveal that the model does not work effectively in the domains that it has not been trained on \cite{lertvittayakumjorn-etal-2020-find,yao2021refining}.
All of these techniques give rise to undesirable model behaviors, requiring debugging.
Another technique, not found in Table~\ref{tab:papers} but suggested in related work \cite{idahl2021towards}, is contaminating input texts in the training data with decoys (i.e., injected artifacts) which could deceive the model into predicting for the wrong reasons. This has been experimented with in the computer vision domain \cite{rieger2020interpretations}, and its use in the \EBHD setting in NLP could be an interesting direction to explore.

\subsection{Workflow} \label{subsec:workflow}
This section describes existing work around the three steps of the \EBHD workflow in Figure~\ref{fig:overview}, i.e., how to generate and present the explanations, how to collect human feedback, and how to update the model using the feedback.
Researchers need to make decisions on these key points harmoniously to create an effective debugging workflow.

\subsubsection{Providing Explanations} 
The main role of explanations here is to 
provide interpretable insights into the model and uncover its potential misbehavior or irrationality, which sometimes cannot be noticed by looking at the model outputs or the evaluation metrics.

\paragraph{Explanation scopes.} 
Basically, there are two main types of explanations that could be provided to feedback providers. Local explanations (L) explain the predictions by the model for individual inputs. 
In contrast, global explanations (G) explain the model overall, independently of any specific inputs.
It can be seen from Table~\ref{tab:papers} that most existing work use local explanations.
One reason for this may be that,
for complex models, 
global explanations can hardly reveal 
details of the models' inner workings 
in a comprehensible way to users. 
So, some bugs are imperceptible in such high-level global explanations and then not corrected by the users.
For example, the debugging framework \textit{FIND}, proposed by \citet{lertvittayakumjorn-etal-2020-find}, uses only global explanations, and it was shown to work more effectively on significant bugs (such as gender bias in abusive language detection) than on less-obvious bugs (such as dataset shift between product types of sentiment analysis on product reviews). 
Otherwise, \citet{ribeiro-etal-2018-semantically} presented adversarial replacement rules as global explanations to reveal the model weaknesses only, without explaining how the whole model worked.

On the other hand, using local explanations has limitations in that it demands a large amount of effort from feedback providers to inspect the explanation of every single example in the training/validation set.
With limited human resources, efficient ways to rank or select examples to explain would be required \cite{idahl2021towards}.
For instance, 
\citet{khanna2019interpreting} and \citet{han2020model} targeted explanations of incorrect predictions in the validation set.
\citet{ribeiro2016should} picked sets of non-redundant local explanations to illustrate the global picture of the model. 
Instead, \citet{teso2019explanatory} leveraged heuristics from active learning to choose unlabeled examples that maximize some informativeness criteria.

Recently, some work in explainable AI considers generating explanations for a group of predictions \cite{johnson2020njm,chan2020subplex} (e.g., for all the false positives of a certain class), thus staying in the middle of the two extreme explanation types (i.e., local and global). 
This kind of explanation is not too fine-grained, yet it can capture some suspicious model behaviors if we target the right group of examples.
So, it would be worth studying in the context of \EBHD 
(to the best of our knowledge, no existing study experiments with it).

\paragraph{Generating explanations.}
To generate explanations in general, there are two important questions we need to answer. 
First, 
which format should the explanations have?
Second, how do we generate the explanations?

For the first question, we see many possible answers in the literature of explainable NLP (e.g., see the survey by \citet{danilevsky-etal-2020-survey}). For instance, \textit{input-based explanations} (so called feature importance explanations) identify parts of the input that are important for the prediction. 
The explanation could be 
a list of importance scores of words in the input, so called \textit{attribution scores} or \textit{relevance scores} \cite{lundberg2017unified,arras-etal-2016-explaining}.
\textit{Example-based explanations} select influential, important, or similar examples from the training set to explain why the model makes a specific prediction \cite{han-etal-2020-explaining,guo2020fastif}.
\textit{Rule-based explanations} provide interpretable decision rules that approximate the prediction process
\cite{ribeiro2018anchors}.
\textit{Adversarial-based explanations} return the smallest changes in the inputs that could change the predictions, revealing the model misbehavior \cite{zhang2020adversarial}.
In most NLP tasks, 
input-based explanations are the most popular approach for explaining predictions \cite{bhatt2020explainable}. This is also the case for \EBHD as most selected studies use input-based explanations \cite{kulesza2009fixing,kulesza2010explanatory,teso2019explanatory,cho2019explanatory}
followed by example-based explanations
\cite{koh2017understanding,khanna2019interpreting,han2020model}. 
Meanwhile, only \citet{ribeiro-etal-2018-semantically} use adversarial-based explanations, whereas \citet{stumpf2009interacting} experiment with input-based, rule-based, and example-based explanations. 

For the second question, there are two ways to generate the explanations: self-explaining methods and post-hoc explanation methods. Some models, e.g., Naive Bayes, logistic regression, and decision trees, are \textit{self-explaining} (SE) \cite{danilevsky-etal-2020-survey}, 
also referred to as transparent \cite{adadi2018peeking} or inherently interpretable \cite{rudin2019stop}.
Local explanations of self-explaining models can be obtained at the same time as predictions, usually from the process of making those predictions, while 
the models themselves can often serve directly as global explanations.
For example, feature importance explanations for a Naive Bayes model can be directly derived from the likelihood terms in the Naive Bayes equation, as done by several papers in Table~\ref{tab:papers} \cite{kulesza2009fixing,smith2020no}.
Also, using attention scores on input as explanations, as done in \cite{cho2019explanatory}, is a self-explaining method because the scores were obtained during the prediction process. 

In contrast, \textit{post-hoc explanation methods} (PH) perform additional steps to extract explanations after the model is trained (for a global explanation) or after the prediction is made (for a local explanation).
If the method is allowed to access model parameters, it may calculate word relevance scores by propagating the output scores back to the input words \cite{arras-etal-2016-explaining} or analyzing the derivative of the output with respect to the input words \cite{smilkov2017smoothgrad,sundararajan2017axiomatic}. 
If the method cannot access the model parameters, it may perturb the input and see how the output changes to estimate the importance of the altered parts of the input \cite{ribeiro2016should,jin2019towards}.
The important words and/or the relevance scores can be presented to the feedback providers in the \EBHD workflow in many forms such as a list of words and their scores \cite{teso2019explanatory,ribeiro2016should}, word clouds \cite{lertvittayakumjorn-etal-2020-find}, and a parse tree \cite{yao2021refining}. 
Meanwhile, the influence functions method, used in \cite{koh2017understanding,zylberajch-etal-2021-hildif}, identifies training examples which influence the prediction by analyzing how the prediction would change if we did not have each training point. This is another post-hoc explanation method as it takes place after prediction. 
It is similar to the other two example-based explanation methods used in \cite{khanna2019interpreting,han2020model}.

\paragraph{Presenting explanations.} 
It is important to carefully design the presentation of explanations, taking into consideration the background knowledge, desires, and limits of the feedback providers.
In the debugging application by \citet{kulesza2009fixing}, lay users were asked to provide feedback to email categorizations predicted by the system. 
The users were allowed to ask several Why questions (inspired by \citet{myers2006answering}) through either the menu bar, or by right-clicking on the object of interest (such as a particular word).
Examples include ``Why will this message be filed to folder A?'', ``Why does word x matter to  folder B?''.
The system then responded by textual explanations (generated using templates), together with visual explanations such as bar plots for some types of questions.
All of these made the interface become more user-friendly.
In \citeyear{kulesza2015principles}, \citeauthor{kulesza2015principles} proposed, as desirable principles, that the presented explanations should be sound (i.e., truthful in describing the underlying model), complete (i.e., not omitting important information about the model), but not overwhelming (i.e., remaining comprehensible).
However, these principles are challenging especially when working on non-interpretable complex models.

\subsubsection{Collecting Feedback}
After seeing explanations, 
humans generally desire to improve the model by giving feedback \cite{smith2020no}.
Some existing work asked humans to confirm or correct machine-computed explanations.
Hence, the form of feedback fairly depends on the form of the explanations,
and in turn this shapes how to update the model too (discussed in section \ref{subsec:update}).
For text classification, 
most \EBHD papers asked humans to decide which words (WO) in the explanation (considered important by the model) are in fact relevant or irrelevant \cite{kulesza2010explanatory,ribeiro2016should,teso2019explanatory}.
Some papers even allowed humans to adjust the word importance scores (WS) \cite{kulesza2009fixing,kulesza2015principles}.
This is analogous to specifying relevancy scores for example-based explanations (ES) in \cite{zylberajch-etal-2021-hildif}.
Meanwhile, feedback at the level of learned features (FE) (i.e., the internal neurons in the model) and learned rules (RU) rather than individual words, was asked in  \cite{lertvittayakumjorn-etal-2020-find} and \cite{ribeiro-etal-2018-semantically}, respectively.
Additionally, humans may be asked to check the predicted labels \cite{kulesza2009fixing,smith2020no} or even the ground truth labels (collectively noted as LB in Table~\ref{tab:papers}) \cite{koh2017understanding,khanna2019interpreting,han2020model}. 
Targeting the table question answering, \citet{cho2019explanatory} asked humans to identify where in the table and the question the model should focus (AT). This is analogous to identifying relevant words to attend for text classification.

It is likely that identifying important parts in the input is sufficient to make the model accomplish simple text classification tasks.
However, this might not be enough for complex tasks which require reasoning. 
Recently, \citet{yao2021refining} asked humans to provide, as feedback, compositional explanations 
to show how the humans would reason (RE) about the models’ failure cases.
An example of the feedback for a hate speech detection is ``Because $X$ is the word dumb, $Y$ is a hateful word, and $X$ is directly before $Y$, the attribution scores of both $X$ and $Y$ as well as the interaction score between $X$ and $Y$ should be increased''.
To acquire richer information like this as
feedback, their framework requires more expertise from the feedback providers.
In the future, it would be interesting to explore how we can collect and utilize other forms of feedback, e.g., natural language feedback \cite{camburu2018snli}, 
new training examples \cite{Fiebrink09ametainstrument}, and other forms of decision rules used by humans \cite{carstens2017using}.

\subsubsection{Updating the Model} \label{subsec:update}
Techniques to incorporate human feedback into the model can be categorized into three approaches.

\paragraph{(1) Directly adjust the model parameters (M).}
When the model is transparent and the explanation displays the model parameters in an intelligible way, humans can directly adjust the parameters based on their judgements. This idea was adopted by \citet{kulesza2009fixing,kulesza2015principles} where humans can adjust a bar chart showing word importance scores, corresponding to the parameters of the underlying Naive Bayes model.
In this special case, steps 2 and 3 in Figure~\ref{fig:overview} are combined into a single step.
Besides, human feedback can be used to modify the model parameters indirectly. 
For example, \citet{smith2020no} increased a word weight in the Naive Bayes model by 20\% for the class that the word supported, according to human feedback, and reduced the weight by 20\% for the opposite class (binary classification). 
This choice gives good results, however, 
it is not clear why and whether 20\% is the best choice here.

Overall, this approach is fast because it does not require model retraining. However, it is important to ensure that the adjustments made by humans generalize well to all examples.
Therefore, the system should update the overall results (e.g., performance metrics, predictions, and explanations) in real time after applying any adjustment, so the humans can investigate the effects and further adjust the model parameters (or undo the adjustments) if necessary. 
This agrees with the correctability principles proposed by \citet{kulesza2015principles} that the system should be actionable and reversible, honor user feedback, and show incremental changes.

\paragraph{(2) Improve the training data (D).} 
We can use human feedback to improve the training data and retrain the model to fix bugs.
This approach includes 
correcting mislabeled training examples \cite{koh2017understanding,han2020model},
assigning noisy labels to unlabeled examples \cite{yao2021refining}, 
removing irrelevant words from input texts \cite{ribeiro2016should},
and creating augmented training examples to reduce the effects of the artifacts \cite{ribeiro-etal-2018-semantically,teso2019explanatory,zylberajch-etal-2021-hildif}.
As this approach modifies the training data only, it is applicable to any model regardless of the model complexity.

\paragraph{(3) Influence the training process (T).}
Another approach is to influence the (re-)training process in a way that the resulting model will behave as the feedback suggests.
This approach could be either model-specific (such as attention supervision) or model-agnostic (such as user co-training).
\citet{cho2019explanatory} used human feedback to supervise attention weights of the model.
Similarly, \citet{yao2021refining} added a loss term to regularize explanations guided by human feedback.
\citet{stumpf2009interacting} proposed 
\textit{(i)} constraint optimization, translating human feedback into constraints governing the training process
and \textit{(ii)} 
user co-training, using feedback as another classifier working together with the main ML model in a semi-supervised learning setting.
\citet{lertvittayakumjorn-etal-2020-find} disabled some learned features deemed irrelevant, 
based on the feedback, and re-trained the model, forcing it to use only the remaining features.
With many techniques available, however, there has not been a study testing which technique is more appropriate for which task, domain, or model architecture. The comparison issue is one of the open problems for \EBHD research (to be discussed in section~\ref{sec:open_problems}).

\subsubsection{Iteration}
The debugging workflow (explain, feedback, and update) can be done iteratively to gradually improve the model 
where the presented explanation changes after the model update. 
This allows humans to fix vital bugs first and finer bugs in later iterations, as reflected in \cite{ribeiro2016should,koh2017understanding} via the performance plots.
However, the interactive process could be susceptible to 
\textit{local decision pitfalls} where local improvements for individual predictions could add up to inferior overall performance \cite{wu2019local}.
So, we need to ensure that the update in the current iteration is generally favorable and does not overwrite the good effects of previous updates.

\subsection{Experimental Setting} \label{subsec:setting}

To conduct experiments, some studies in Table~\ref{tab:papers} selected human participants (SP) to be their feedback providers.
The selected participants could be people without ML/NLP knowledge \cite{kulesza2010explanatory,kulesza2015principles} or with ML/NLP knowledge \cite{ribeiro-etal-2018-semantically,zylberajch-etal-2021-hildif} depending on the study objectives and the complexity of the feedback process.
Early work even conducted experiments with the participants in-person \cite{stumpf2009interacting,kulesza2009fixing,kulesza2015principles}.
Although this limited the number of participants (to less than 100), the researchers could closely observe their behaviors and gain some insights concerning human-computer interaction.

By contrast, some used a crowdsourcing platform, Amazon Mechanical Turk\footnote{\url{https://www.mturk.com/}} in particular, to collect human feedback for debugging the models.
Crowdsourcing (CS) enables researchers to conduct experiments at a large scale; however, the quality of human responses could be varying. So, it is important to ensure some quality control such as specifying required qualifications \cite{smith2020no}, using multiple annotations per question \cite{lertvittayakumjorn-etal-2020-find}, having a training phase for participants, and setting up some obvious questions to check if the participants are paying attention to the tasks
\cite{egelman2014crowdsourcing}.

Finally,  simulation (SM), without real humans involved but using oracles as human feedback instead, has also been considered (for the purpose of testing the \EBHD framework only). 
For example, \citet{teso2019explanatory} set 20\% of input words as relevant using feature selection.
These were used to respond to 
post-hoc explanations, i.e., top $k$ words selected by LIME.
\citet{koh2017understanding} simulated mislabeled examples by flipping the labels of a random 10\% of the training data.
So, when the explanation showed suspicious training examples, the true labels could be used to provide feedback.
Compared to the other settings,
simulation is faster and cheaper, yet its results may not reflect the effectiveness of the framework when deployed with real humans. Naturally, human feedback is sometimes inaccurate and noisy, and humans could also be interrupted or frustrated while providing feedback \cite{amershi2014power}. These factors, discussed in detail in the next section, cannot be thoroughly studied in only simulated experiments.

\section{Research on Human Factors} \label{sec:human_factor}
Though the major goal of \EBHD is to improve models, we cannot disregard the effect on feedback providers of the debugging workflow. 
In this section, we compile findings 
concerning how explanations and feedback could affect the humans, discussed 
along five dimensions: model understanding,  willingness, trust, frustration, and expectation.
Although some of the findings were not derived in NLP settings, we believe that they are generalizable and worth discussing in the context of \EBHD.

\subsection{Model Understanding}
So far, we have used explanations as means to help humans understand models and conduct informed debugging.
Hence, it is important to verify, at least preliminarily, that the explanations help feedback providers form an accurate understanding of how the models work. This is an important prerequisite towards successful debugging.

Existing studies have found that some explanation forms are more conducive to developing model understanding in humans than others. 
\citet{stumpf2009interacting} found that rule-based and keyword-based explanations were easier to understand than similarity-based explanations (i.e., explaining by similar examples in the training data).
Also, they found that some users did not understand why the absence of some words could make the model become more certain about its predictions. 
\citet{lim2009and}
found that explaining why the system behaved and did not behave in a certain way resulted in good user's understanding of the system, though the former way of explanation (why) was more effective than the latter (why not).
\citet{cheng2019explaining}
reported that interactive explanations could improve users’ comprehension on the model better than static explanations, although the interactive way took more time. 
In addition, 
revealing inner workings of the model could further help understanding; however, it introduced additional cognitive workload that might 
make participants doubt whether they really understood the model well.

\subsection{Willingness}
We would like humans to provide feedback for improving models, but do humans naturally want to?
Prior to the emerging of \EBHD, studies found that humans are not willing to be constantly asked about labels of examples as if they were just simple oracles \cite{cakmak2010designing,guillory2011simultaneous}.
Rather, they want to provide more than just data labels after being given explanations \cite{amershi2014power,smith2020no}.
By collecting free-form feedback from users, \citet{stumpf2009interacting} and \citet{ghai2021explainable}
discovered various feedback types. The most prominent ones include removing-adding features (words), tuning weights, and leveraging feature combinations.
\citet{stumpf2009interacting} further analyzed categories of background knowledge underlying the feedback and found, in their experiment, that it was mainly based on commonsense knowledge and English language knowledge. 
Such knowledge may not be efficiently injected into the model if we exploit human feedback which contains only labels.
This agrees with some participants, in \cite{smith2020no}, who described their feedback as inadequate when they could only confirm or correct predicted labels.

Although human feedback beyond labels contains helpful information, it is naturally neither complete nor precise. \citet{ghai2021explainable}
observed that human feedback usually focuses on a few features that are most different from human expectation, ignoring the others.
Also, they found that humans, especially lay people, are not good at correcting model
explanations quantitatively (e.g., adjusting weights). 
This is consistent with the findings of
\citet{miller2019explanation}
that human explanations are selective (in a biased 
way) and rarely refer to probabilities but express causal relationships instead.

\subsection{Trust}
Trust (as well as frustration and expectation, discussed next) is an important issue when the system end users are feedback providers in the \EBHD framework.
It has been discussed widely that explanations engender human trust in AI systems \cite{pu2006trust,lipton2018mythos,toreini2020relationship}.
This trust may be misplaced at times. 
Showing more detailed explanations can cause users to over rely on the system, leading to misuse where users agree with incorrect system predictions \cite{stumpf2016explanations}.
Moreover, some users may over trust the explanations (without fully understanding them) only because the tools generating them are publicly available, widely used, and showing appealing visualizations 
\cite{kaur2020interpreting}.

However, recent research reported that explanations do not necessarily increase trust and reliance.
\citet{cheng2019explaining}
found that, even though explanations help users comprehend systems, they cannot increase human trust in using the systems in high-stakes applications involving lots of qualitative factors, such as graduate school admissions.    
\citet{smith2020no} reported that explanations of low-quality models decrease trust and system acceptance as they reveal model weaknesses to the users.
According to \citet{schramowski2020making}, despite correct predictions, the trust still drops if the users see from the explanations that the model relies on the wrong reasons. 
These studies go along with a perspective by \citet{zhang2020effect} that explanations should help calibrate user perceptions to the model quality, signaling whether the users should trust or distrust the AI.
Although, in some cases, explanations successfully warned users of faulty models \cite{ribeiro2016should}, this is not easy when the model flaws are not obvious \cite{zhang2020effect,lertvittayakumjorn-toni-2019-human}. 

Besides explanations, the effect of feedback on human trust is quite inconclusive according to some (but fewer) studies.
On one hand, 
\citet{smith2020no} found that, after lay humans see explanations of low-quality models and lose their trust, the ability to provide feedback makes human trust and acceptance rally, remedying the situation.
In contrast, \citet{honeycutt2020soliciting} reported that providing feedback decreases human trust in the system as well as their perception of system accuracy no matter whether the system truly improves after being updated or not.

\subsection{Frustration}
Working with explanations can cause frustration 
sometimes.
Following the discussion on trust, explanations of poor models increase user frustration (as they reveal model flaws), whereas the ability to provide feedback reduces frustration. Hence, in general situations, the most frustrating condition is showing explanations to the users without allowing them to give feedback \cite{smith2020no}. 

Another cause of frustration is the risk of detailed explanations overloading users \cite{narayanan2018humans}.
This is especially a crucial issue for inherently interpretable models where all the internal workings can be exposed to the users.
Though presenting all the details is comprehensive and faithful, it could create
barriers for lay users \cite{gershon1998visualization}.
In fact, even ML experts may feel frustrated if they need to understand a decision tree with a depth of ten or more.
\citet{poursabzi2018manipulating} found that showing all the model internals undermined users' ability to detect flaws in the model, likely due to information overload. So, they suggested that model internals should be revealed only when the users request to see them.

\subsection{Expectation}
\citet{smith2020no} observed that some participants expected the model to improve after the session where they interacted with the model, regardless of whether they saw explanations or gave feedback during the interaction session. 
\EBHD should manage these expectations properly. 
For instance, the system should report changes or improvements to users after the model gets updated.
It would be better if the changes can be seen incrementally in real time \cite{kulesza2015principles}.

\subsection{Summary}
Based on the findings on human factors reviewed in this section, we summarize suggestions for effective \EBHD as follows.

\paragraph{Feedback providers.} Buggy models usually lead to implausible explanations, adversely affecting human trust in the system.
Also, it is not yet clear whether giving feedback increases or decreases human trust.
So, it is safer to let the developers or domain experts in the team (rather than end users) be the feedback providers.
For some kinds of bugs, however, feedback from end users is essential for improving the model. 
To maintain their trust, we may collect their feedback implicitly (e.g., by inferring from their interactions with the system after showing them the explanations \cite{honeycutt2020soliciting}) or collect the feedback without telling them that the explanations are of the production system (e.g., by asking them to answer a separate survey). 
All in all, we need different strategies to collect feedback from different stakeholders.

\paragraph{Explanations.} We should avoid using forms of explanations which are difficult to understand, such as similar training examples and absence of some keywords in inputs, unless the humans are already trained to interpret them.
Also, too much information should be avoided as it could overload the humans; instead, humans should be allowed to request more information if they are interested,
e.g., by using interactive explanations \cite{dejl2021argflow}.

\paragraph{Feedback.} Given that human feedback is not always complete, correct, or accurate,
\EBHD should use it with care, e.g., by relying on collective feedback rather than individual feedback and allowing feedback providers to verify and modify their feedback before applying it to update the model.

\paragraph{Update.} Humans, especially lay people, usually expect the model to improve over time after they give 
feedback. So, the system should display improvements after the model gets updated.
Where possible, showing the changes incrementally 
in real time is preferred, as the feedback providers can check if their feedback works as expected or not.

\section{Open Problems} \label{sec:open_problems}
This section lists potential research directions and open problems for \EBHD of NLP models.

\subsection{Beyond English Text Classification}

All papers in Table~\ref{tab:papers} conducted experiments only on English datasets. 
We acknowledge that qualitatively analyzing explanations and feedback in languages at which one is not fluent is not easy, not to mention recruiting human subjects who know the languages.   
However, we hope that, with more multilingual data publicly available \cite{2020HuggingFace-datasets} and growing awareness in the NLP community \cite{bender2019rule}, there will be more \EBHD studies targeting other languages in the near future.

Also, most existing \EBHD works target  
text classifiers. It would be interesting to conduct more \EBHD work for other NLP tasks such as reading comprehension, question answering, and natural language inference (NLI), to see whether existing techniques still work effectively. 
Shifting to other tasks requires an understanding of specific bug characteristics in those tasks.
For instance, unlike bugs in text classification which are usually due to word artifacts, bugs in NLI concern syntactic heuristics between premises and hypotheses \cite{mccoy-etal-2019-right}. Thus, giving human feedback  
at word level may not be helpful, and more advanced methods may be needed.

\subsection{Tackling More Challenging Bugs}
\citet{singh2020neurips} remarked that the evaluation setup of existing \EBHD work is often too easy or unrealistic. For example, bugs are obvious artifacts which could be removed using simple text pre-processing (e.g., removing punctuation and redacting named entities). 
Hence, it is not clear how powerful such \EBHD frameworks are when dealing with real-world  bugs. 
If bugs are not dominant and happen less often, global explanations may be too coarse-grained to capture them while 
many local explanations may be needed to spot a few appearances of the bugs, leading to inefficiency.
As reported by \citet{smith2020no},
feedback results in minor improvements when the model is already reasonably good. 

Other open problems, whose solutions may help deal with  
challenging bugs, include the following.
First, different people may give different feedback for the same explanation. As raised by \citet{ghai2021explainable}, how can we integrate their feedback to get robust signals for model update? How should we deal with conflicts among feedback and training examples \cite{carstens2017using}?
Second, confirming or removing what the model has learned is easier than injecting, into the model, new knowledge (which may not even be apparent in the explanations). How can we use human feedback to inject new knowledge, especially when the model is not transparent?
Lastly, \EBHD techniques have been proposed for tabular data and image data \cite{shao2020towards,ghai2021explainable,popordanoska2020machine}. Can we adapt or transfer them across modalities to deal with NLP tasks?

\subsection{Analyzing and Enhancing Efficiency}
Most selected studies focus on improving correctness of the model (e.g., by expecting a higher F1 or a lower bias after debugging).
However, 
only some of them discuss efficiency of the proposed frameworks.
In general, we can analyze the efficiency of an \EBHD framework by looking at the efficiency of each main step in Figure~\ref{fig:overview}.
Step 1 generates the explanations, so its efficiency depends on the explanation method used and, in the case of local explanation methods, the number of local explanations needed.
Step 2 lets humans give feedback, so its efficiency concerns the amount of time they spend to understand the explanations and to produce the feedback.
Step 3 updates the model using the feedback, so its efficiency relates to the time used for processing the feedback and retraining the model (if needed).
Existing work mainly reported efficiency of step 1 or step 2.
For instance, approaches using example-based explanations 
measured the improved performance with respect to the number of explanations computed (step 1) \cite{koh2017understanding,khanna2019interpreting,han2020model}. 
\citet{kulesza2015principles} compared the improved F1 of \EBHD with the F1 of instance labelling given the same amount of time for humans to perform the task (step 2).
Conversely, \citet{yao2021refining} compared the time humans need to do \EBHD versus instance labelling in order to achieve the equivalent degree of correctness improvement (step 2).

None of the selected studies considered the efficiency of the three steps altogether. In fact, the efficiency of step 1 and 3 is important especially for black box models where the cost of post-hoc explanation generation and model retraining is not negligible. 
It is even more crucial for iterative or responsive \EBHD.
Thus, analyzing and enhancing efficiency of \EBHD frameworks (for both  machine  and  human sides) require further research.

\subsection{Reliable Comparison across Papers}
 
User studies 
are naturally difficult to replicate as they are inevitably affected by choices of user interfaces, phrasing, population, incentives, etc. \cite{singh2020neurips}.
Further, research in ML rarely 
adopts practices from the human-computer interaction community \cite{abdul2018trends}, limiting the possibility to compare across studies.
Hence, most existing work only considers model performance before and after debugging or compares the results among several configurations of a single proposed framework.
This leads to little knowledge 
about which explanation types or feedback mechanisms are more effective across several settings.
Thus, one promising research direction would be proposing a standard setup or a benchmark for
evaluating and comparing \EBHD frameworks 
reliably across different settings. 

\subsection{Towards Deployment}
So far, we have not seen \EBHD research widely deployed in applications,
probably due to its difficulty to set up the debugging aspects outside a research environment.
One way to promote adoption of \EBHD is to integrate \EBHD frameworks into available visualization systems such as the Language Interpretability Tool (LIT) \cite{tenney2020language}, allowing users to provide feedback to the model after seeing explanations and supporting experimentation.
Also, to move towards deployment, it is important to follow human-AI interaction guidelines \cite{amershi2019guidelines} and evaluate \EBHD with potential end users, not just via simulation or crowdsourcing, since human factors play an important role in real situations \cite{amershi2014power}.

\section{Conclusion} \label{sec:conclusion}
We presented a general framework of \ebhd (\EBHD) of NLP models and analyzed existing work in relation to the components of this framework to illustrate the 
state-of-the-art in the field. 
Furthermore, we summarized findings on human factors with respect to \EBHD, suggested design practices accordingly, and identified open problems for future studies.
As \EBHD is still an ongoing research topic, we hope that our survey will be helpful for guiding interested researchers 
and for examining future \EBHD papers.

\section*{Acknowledgments}
We would like to thank Marco Baroni (the action editor) and anonymous reviewers for very helpful comments. 
Also, we thank Brian Roark and Cindy Robinson for their technical support concerning the submission system.
Besides, the first author wishes to thank the support from Anandamahidol Foundation, Thailand.

\bibliography{anthology,survey}
\bibliographystyle{acl_natbib}

\end{document}